# Computational identification of regulatory statements in EU legislation

Gijs Jan Brandsma[a] Jens Blom-Hansen[b] Christiaan Meijer[c] Kody Moodley[1,c,d]

[a] *Institute of Management Research, Radboud University, Nijmegen, The Netherlands*
[b] *Aarhus University, Aarhus, Denmark*
[c] *The Netherlands eScience Center, Amsterdam, The Netherlands*
[d] *ORCID: 0000-0001-5666-1658*

[gijsjan.brandsma@ru.nl](gijsjan.brandsma@ru.nl), [jbh@ps.au.dk](jbh@ps.au.dk), [{k.moodley;c.meijer}@esciencecenter.nl]({k.moodley;c.meijer}@esciencecenter.nl)

## ABSTRACT

Identifying regulatory statements in legislation is useful for developing metrics to measure the regulatory density and strictness of legislation, and to study how these vary over time and across policy areas. A computational method is valuable for scaling the identification of such statements from a growing body of EU legislation, constituting approximately 180,000 published legal acts between 1952 and 2023. Past work on extraction of these statements varies in the permissiveness of their definitions for what constitutes a regulatory statement. In this work, we provide a specific definition for our purposes based on the institutional grammar tool (IGT). We then develop and compare two contrasting approaches for automatically identifying such statements in EU legislation, one based on dependency parsing, and the other on a transformer-based machine learning model. We found that both approaches performed similarly well with accuracy scores of 0.80 and 0.84 respectively and a Krippendorff alpha coefficient of 0.58. The high accuracies and not exceedingly high agreement suggests that accuracy could be further increased by combining the strengths of both approaches.

## 1 Introduction

In the period from 1952 to 2023 there have been approximately 180,000 EU legislative documents published that are publicly available for digital viewing or download in the English language. The documents vary according to the legal matters they address e.g., taxation, climate policy or agriculture and generally consist of multiple provisions or articles. To be able to study this large and growing body of legislation, computational methods are becoming increasingly necessary. One specific computational task is the automatic identification of regulatory statements in legal text. This task is a necessary step for statistical analyses measuring the density and strictness of legislation. Both these tasks have many concrete applications in legal practice, education and scholarship. For example, law firms may use such technology to quickly identify relevant regulatory risks for clients; scholars and students may use it to study the evolution of regulatory frameworks in different industries (e.g., health care and fintech); organizations responsible for drafting legislation can make use of it to assess effectiveness of past regulations and inform future legislative drafting.

Methods for identifying regulatory sentences and legal obligations in text use the presence deontic expressions (keywords such as "*must*" and "*shall*") in the text to help classify statements. However, such keywords alone do not necessarily indicate that a sentence is regulatory. For example, sentences such as "*It shall apply from 23 November 2016.*" and "*This Decision shall enter into force on the date of its adoption.*", frequently appear in EU legislative decisions and contain the deontic expression "*shall*", but neither indicates a legal rule intended to regulate a specific legal entity.

Deontic sentence classification (DSC) in legal text is a closely related area which seeks to computationally classify deontic statements in legal text into different categories based on their normative nature (e.g., into obligations, permissions and prohibitions). While deontic logic [9, 10] provides a way to represent and reason with legal statements expressing obligations, more recent efforts have used Natural Language Processing and Machine Learning to classify deontic modalities in text [8, 11, 17, 18, 35]. The main characteristic of DSC which limits its applicability for our goals of identifying regulatory statements in text is the permissiveness in its definition. For example, consider the following text "*Citizens must separate their recyclables. This must be done by them under the following conditions: (i) before disposing of trash, and (ii) if (i) is not met, one is subject to a fine*". There are two sentences in this text and DSC recognizes both as regulatory statements expressing legal obligations, permissions or prohibitions. However, in our definition of regulatory statement, we consider only the first sentence as indicating the core regulation and the second as providing the preconditions that need to be met for the rule in the first sentence to be satisfied.

[1] Corresponding author

We give a more detailed account of our definition of regulatory statement, which we base on the *Institutional Grammar Tool* (IGT) framework developed by Crawford and Ostrom [6, 25], in Section 3.1. Furthermore, prior work on *binary* classification of regulatory statements has not yielded models that have been trained and evaluated on EU legislation that is *agnostic of policy area and time-period* (see Section 2). Past work has also not compared grammatical dependency analysis and machine learning approaches for this task in detail. Not just in terms of performance, but also in terms of characterizing model predictions and highlighting complementary strengths and weaknesses.

In this paper we address these open issues by presenting, evaluating and comparing two approaches for computationally identifying regulatory statements in EU legislation. We frame the task as a binary (short) text classification problem where the goal is to classify a given sentence from EU legislation as regulatory or non-regulatory (capturing a legal rule targeting a specific legal entity, or not). The first approach we present analyzes the grammatical dependencies in the sentences to verify if it meets all the criteria in our IGT definition for being a regulatory statement. The second approach uses feature-based transfer learning from a pretrained transformer model known as *LEGAL-BERT* [4]. To the best of our knowledge, this latter approach involved the creation of the largest corpus to date of human-labelled sentences for binary text classification of regulatory statements (7,200 sentences). We also compare the two approaches according to standard classification performance metrics and agreement. We utilized *eXplainable AI (XAI)* software called *Deep Insight And Neural Network Analysis* or *DIANNA* for short [30] to give preliminary insight into LEGAL-BERT classification behavior in a sample of our sentences.

The novel contributions of this paper are:
1. Two high-performing binary classification approaches for regulatory statements in general EU legislation (agnostic of policy area and time-period).
2. A performance and predictions comparison of a grammatical dependency vs. a feature-based transfer learning approach for this task.
3. XAI analysis of the feature-based learning approach to enhance the interpretability of classifications by humans.

The rest of this paper is organized as follows: in Section 2 we give a brief overview of closely related work in DSC and text classification for EU law. In Section 3 we describe in detail our definition of regulatory statement, our two approaches for regulatory statement classification and the setup of our XAI study. Section 4 presents the results and comparison for our two approaches, and the XAI analysis. Section 5 discusses some challenges we encountered in the study and some caveats to keep in mind when interpreting the results. Section 6 gives some details of our code, data and analysis files. We conclude with a summary and directions for future work in Section 7.

## 2 Related Work

The most closely related problem to ours in the literature is DSC which aims to automatically classify statements in legal text according to their deontic types [2, 15, 17, 24]. Several class labels are proposed including obligations, permissions, prohibitions and constitutive statements. This problem has been explored in other types of legal text as well, not just legislation, but also contracts [11, 15, 32].

Dragoni et al. make use of Natural Language Processing, ontologies and logic-based dependency analysis of sentences in the Australian Telecommunications Consumer Protection Code to perform DSC [8]. Although they go a bit further than DSC by decomposing the structure of the sentences to model their fine-grained deontic or normative structure. O'Neill et al. appear to be the first to apply neural network classifiers on EU legislative sentences for DSC [24]. Their data also comprises EU legislation (along with UK legislation), but they focus the DSC task on financial regulations specifically using the same target classes of obligations, permissions and prohibitions. They also make use of word embeddings, particularly *word2vec* [23] and *GloVe* [28], as feature representations for input into various classifiers. They tried both non-neural network and neural network architectures with the neural networks unsurprisingly outperforming the former, Bidirectional LSTMs [13] proving to be the best performing overall.

More recently, Liga and Palmirani [18] applied transfer learning for DSC by fine-tuning three existing pretrained models: *Bidirectional Encoder Representations from Transformers (BERT)* [8], and subsequent BERT variations, DistilBERT [33] and LEGAL-BERT. They apply DSC to sentences in General Data Protection Regulation (GDPR) provisions to determine whether they contain a legal rule or not, whether they contain deontic information or not, and further subclassification either as an obligation, permission, or constitutive rule. The data was represented in LegalXML documents annotated using the Akoma Ntoso [26] and LegalRuleML [1] formats.

DSC has also been applied in contracts. Sancheti et al. [32] created a benchmark for DSC in contracts called *LEXDEMOD*. They evaluated this benchmark on tasks concerning the prediction of all deontic types expressed in a sentence with respect to an agent mentioned in the sentence. The researchers also employed the use of pretrained language models including *RoBERTa* [20] which is an optimized version of BERT. The rise of contemporary Large Language Models (LLMs) such as *GPT-3* also raised the question of the applicability of such models for DSC and rule classification. Liga and Robaldo investigate this approach by fine-tuning GPT-3 for this task [37]. The data for fine-tuning was the same GDPR data used by Liga and Palmirani [18]. They also compared the performance of their approach to those of Liga and Palmirani. The GPT-3 model obtained an accuracy above 0.9 and outperformed the previous approach by approximately 10%.

While previous approaches have been relatively successful in terms of accuracy, there are still some under-explored areas. Firstly, high performing *binary* classification models for regulatory statements in EU legislation *agnostic of policy area*

*and time-period* is lacking. Secondly, there are no mature comparisons (not just of performance but of prediction behavior) between grammatical dependency and transformer model approaches. While DSC is related to our task of identifying regulatory statements in legal text, our definition of regulatory is based solely on the IGT. While IGT analysis and DSC have similar goals, they differ in how they define regulatory statements. More details about this difference and our IGT-based definition of regulatory statement are provided in Section 3.1.

## 3 Methodology

In this section, we describe our approaches for regulatory sentence classification and the evaluation set up. We start by defining what we mean by a regulatory statement.

### 3.1 Defining regulatory statements

As briefly mentioned in Section 1, not all deontic sentences in a legal text are relevant for our study. For example, to develop metrics to measure density of EU legislation one could, as a starting point, count the number of regulatory sentences in the text. Therefore, while some sentences in legislation may contain deontic expressions and may give important preconditions and contextual information about a legal rule, we may not necessarily want to count them as *distinct* legal rules towards our density metric. The separation of recyclables example in Section 1 demonstrates this point. We may define a metric for legislative density which only counts sentences that specify the core regulation and other sentences which explain further preconditions and context as irrelevant (as one can imagine, the number of sentences used to detail preconditions and context for legal rules can vary widely and, hence, skew measurements).

In this work we are primarily interested in the subset of sentences in EU law which mention specific legal obligations to be carried out or upheld by a specific legal entity. I.e., we are focused on *whether a particular sentence specifies a regulation* and *who is being regulated*. Whether the sentence fully describes the regulation, and its preconditions or criteria, is not critical. We make use of the IGT framework to define such statements. The IGT specifies a framework for analyzing statements made in text concerning institutions (and is therefore also applicable to policy and legal texts). The main analytical components for IGT are:

1. *Attributes (A)*: The actors or individuals to whom a statement applies
2. *Deontic (D)*: The deontic keyword in the statement i.e. "*must*" or "*shall*"
3. *Aim (I)*: The action or outcome that the statement addresses
4. *Condition (C)*: The circumstances under which the statement applies
5. *Or-else (O)*: The sanctions or consequences for non-compliance

This is sometimes known as the ADICO approach to institutional statement analysis. For example, in the sentence: "*Citizens must separate their recyclables before disposing of trash, or else face a fine.*" we have "*Citizens*" as the attribute, "*must*" as the deontic, "*separation of recyclables*" as the aim, "*before disposing of trash*" as the condition, and "*face a fine*" as the or-else clause. In our study, we observed that EU legislation often spreads the ADICO components of a regulation across multiple sentences. That is, sometimes the ADI part is essentially stating what the legal obligation is (often in one sentence). And the CO part further elaborates on preconditions and other criteria concerning the legal obligation (often in subsequent sentences). For example, we can partition the above example sentence into: "*Citizens must separate their recyclables.*" and "*This must be done by them under the following conditions: (i) before disposing of trash, and (ii) if (i) is not met, one is subject to a fine*". We can observe that the first sentence captures the core regulation while the second specifies the triggering preconditions for the rule to be satisfied. In summary, our definition for a regulatory sentence in EU legislation is a sentence which:

1. explicitly mentions at least the ADI components of the IGT framework, **and**
2. expresses the attribute (A) component as an *agent noun* where an agent noun is a noun that describes a person or thing that performs an action. The concept of agent noun subsumes proper nouns computationally identifiable through part of speech (POS) tagging e.g., "*United Kingdom*". But it also includes the class of nouns for which the classification problem has not yet been universally defined e.g., "*Citizen*", "*Cyclist*", "*Councilor*" etc.

### 3.2 Data selection and preparation

The approaches and evaluations presented in this paper were developed as a step towards our main goal, which is to examine the density and distribution of regulatory statements across time and policy area in EU law. To make our classifiers potentially applicable to the entire body of EU law, we therefore selected all EU legislation digitally accessible from official online portals between 1970 and 2022 as our evaluation corpus. 2023 and 2024 legislation were not included because the year had not expired at the time of this evaluation. Similarly, acts pertaining to the years prior to 1970 were not publicly available in the same online portals as for the other years.

We accessed the full texts of the legislation from the EURLEX[2] portal and the metadata (identifiers, dates of adoption, policy area code, legal form, addressee etc.) for the regulations from the CELLAR[3] RESTful API service accessible through its SPARQL interface[4]. This data set comprised roughly 120,000 legislative documents. The mean document length was 4,000 words with many exceedingly long and short outlier documents (standard deviation and median values were approximately 9,000 and 1,000 words respectively).

---

[2] http://eur-lex.europa.eu
[3] http://data.europa.eu/data/datasets/sparql-cellar-of-the-publications-office
[4] http://publications.europa.eu/webapi/rdf/sparql

To evaluate our text classification algorithms, which will be presented in Sections 3.3 and 3.4 respectively, we extracted candidate sentences from the dataset. We did this using the following approach. We first preprocessed the text in all the documents to eliminate content that would not contain potentially regulatory statements. These are elements such as the explanation of legal base, introductory recitals, appendices, images, graphs, tables etcetera. Fortunately, EU legislative documents have a consistent structure. There are always key phrases to indicate the start and end of the main regulatory section or portion of the document: the articles. Phrases such as "HAS ADOPTED THIS REGULATION", "HAVE ADOPTED THIS DECISION" and "HAS ADOPTED THE FOLLOWING DIRECTIVE" always immediately precede the part of the text which contains all regulatory statements. Similarly, the end of the regulatory section of the document (before the appendices and supplementary information) is always signaled by phrases such as "*Done at Brussels*", "*Done at Luxembourg*", "*Done at Strasbourg*" and "*Done at Frankfurt*". A dictionary of such start and end phrases was compiled by the first two authors of this article: an associate professor of public administration and a professor of political science, both specialized in EU public policy. Once we captured the regulatory sections of the documents, we tokenized the text in these sections into individual sentences. The sentences were then filtered to only include those which contain at least one deontic phrase of either "*shall*" or "*must*" (note this approach would also capture the negated deontic phrases "*shall not*" and "*must not*"). We were left with approximately 0.6 million candidate sentences in this corpus after this data preparation process.

### 3.3 Approach 1 - Grammatical Dependency Analysis

The first approach we employed for classifying a sentence as regulatory or not was based on analyzing its grammatical dependencies using neural network-based dependency parsers [12] as implemented in software libraries such as SpaCy[5]. The specific version of SpaCy used was 3.7.2 and we made use of the default English model "en_core_web_sm" version 3.7.1. Initially, we randomly selected one hundred sentences from our prepared sentence corpus described in Section 3.2. We asked the first two authors of this article mentioned in Section 3.2 to label each sentence as regulatory (indicated with a 1) or not (indicated by a 0), in line with the IGT framework. The data was recorded in a two-columned CSV (comma-separated values) file, the first column containing the sentence and the second column containing the classified result which serves as our ground truth. We then manually analyzed the dependency parse trees of each of these sentences to establish a link between the grammatical dependencies in the sentences and the relevant ADICO components mentioned in Section 3.1. I.e., the goal of this task was to examine what patterns in the dependency parse tree of the sentences are induced by our ADICO definition of regulatory statement.

We found the following conditions to be sufficient in most cases (88 out of 100 sentences) for classifying a sentence as regulatory. There must be at least one lexical verb (partially indicating the aim or I component of the IGT framework) in the sentence which satisfies the following criteria:

1. The lexical verb must have a grammatical dependency (relation) to an auxiliary verb which constitutes a deontic phrase either "*must*" or "*shall*"[6].
2. There must be a path (initially starting in either a backwards or forwards direction) through the parse tree from this lexical verb to an agent noun. Direction in this context does not refer to the direction of the arrows (dependencies) in the parse tree, but rather to sequence in the sentence. For example, from the "*separate*" verb in Figure 1, there are two paths backwards, one terminating at the "*must*" token and one terminating at the "*Citizens*" token.

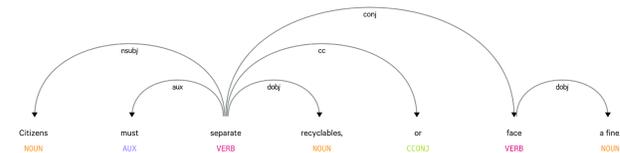

**Figure 1.** Visualization of the grammatical dependencies in a potentially regulatory sentence.

For the second criterion, a path *backwards* indicates we are trying to identify the sentence's main subject. For regulatory statements, we found that this subject often represents the main attribute or A component of the IGT framework when the sentence is written in active voice. For passive voice sentences (which are regulatory), we found that a path initially beginning *forwards* from the lexical verb is the most likely path to the attribute (see Figure 2 for an example).

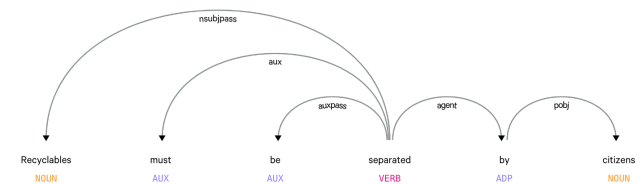

**Figure 2.** Visualization of dependencies in a potentially regulatory sentence written in the passive voice.

We found the identification of agent nouns to be a non-trivial task in English and in other languages [14]. Modern computational POS taggers and dependency parsers can quite accurately identify proper nouns [22]. However, proper nouns are just one specific type of agent noun (which can indicate the attribute of a regulation in a sentence) e.g., "*EU Member States*",

---

[5] http://spacy.io

[6] Negations, for example "must not" and "shall not", are automatically taken care of by the rule

"*France*", "*UK*" etc. Agent nouns that are not proper nouns can also serve as attributes for a regulatory statement. E.g., "*Citizens*", "*Workers*", "*Councilors*" etc.

Unfortunately, there have been no mature attempts to include such terms as distinct class labels in modern POS tagging models. We chose to deal with this by classifying such words as agent nouns through dictionary look up. Specifically, we used ConceptNet [34] to identify such words. ConceptNet is an open multilingual knowledge graph of terms and phrases containing expert and crowd-sourced metadata about such terms and their linguistic relationships with each other. Examples of such relations could be semantic similarity, synonyms, etymological relations, word senses etc. Examples of other metadata for terms in ConceptNet include semantic categories and hyponyms e.g., "*man*" is a hyponym for the ConceptNet class "*agent*" (which maps to what we define as agent nouns). For each noun in a sentence, we determined its status as an agent noun by the absence or possession of a hyponymic relation to the "*agent*" class in ConceptNet.

As stated before, this approach yielded high classification accuracy for the initial sample of 100 sentences (88 out of the 100 sentences were correctly classified). We analyzed the false positives to try to characterize the limitation of our algorithm for these cases. We observed that some sentences appear to satisfy all the criteria mentioned above for a regulatory statement, but the attribute identified does not represent a valid legal entity (see Figure 3). In that example, while "*authorized operators*" is a valid agent noun, it does not serve as an attribute (the legal entity being regulated). Indeed, such an entity is not mentioned in the sentence and therefore according to the IGT framework the statement cannot be classed as a regulatory statement (even though our algorithm would classify it as such).

To address this issue our dependency rules could be updated to account for specific dependency types. For example, if the sentence in Figure 3 is updated to "*Recyclables must be separated by authorized operators*" then "*authorized operators*" becomes the attribute of the sentence, and the dependency parser indicates this though the "*agent*" universal dependency type (which would replace the "*prep*" dependency in Figure 3.

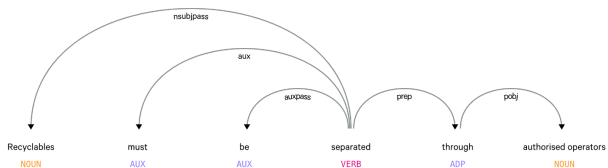

**Figure 3.** Visualization of dependencies in a potentially regulatory sentence: false positive example.

We found that most false negative cases were due to uncommon linguistic variation (phrasing). That is, our algorithm overlooked the attribute who is being regulated in many of the false negative cases (successfully identifying the attribute is a prerequisite for classifying a statement as regulatory according to this approach).

Consider the following example statement in the decision with CELEX identifier 32020R0723: "*In this case, the privileges of the holder shall be limited by the competent authority to performing flight instruction and testing for initial issue of type ratings, the supervision of initial line flying by the operators' pilots, delivery or ferry flights, initial line flying, flight demonstrations or test flights, as appropriate to the tasks foreseen under this paragraph.*" (the dependency diagram is too large to fit into an image on this paper). Here, our parser classifies the statement as written in passive voice which makes sense given the phrase "*…shall be limited by…*" and then tries to follow a forward path from the verb "*limited*" through the statement to reach the attribute. It reaches "*the competent authority*" (which does not appear in ConceptNET and, therefore, the sentence is misclassified as non-regulatory). However, the problem here is that "*the competent authority*" does not refer to the intended target of the legal rule. The "*the holder*" phrase does, but it is not recognized because the parser detects passive voice and, therefore, expects the attribute to lie on a *forward* direction path from the lexical verb and deontic.

### 3.4 Approach 2 – Feature extraction from Legal-BERT

For the transfer-learning approach, we trained a shallow binary text classification model on top of features extracted with pretrained BERT models. We identified three models that could serve as a feature-extraction model. LEGAL-BERT was a natural candidate based on its transformers architecture and it was trained on English legal corpora (which incidentally includes EU legislation, but only up to the year 2020) of 0.45 million documents. The model is also well documented and publicly accessible[7]. A second adequate feature extraction model we identified was InLegal-BERT [27] which is an extension of LEGAL-BERT, trained on an additional corpus of 5.4 million Indian legal documents in the English language. We also included BERT [7] as a baseline feature extraction model. We used XGBoost [5] to train a binary classification model on the extracted features. See Figure 4 for a schematic.

---

[7] http://huggingface.co/nlpaueb/legal-bert-base-uncased

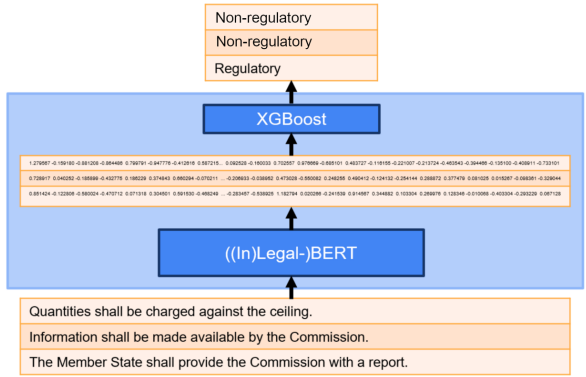

**Figure 4.** Transfer learning model setup. Features are extracted from each sentence in our dataset, after which a shallow classifier was trained using the extracted features.

We requested two human annotators (the researchers mentioned in Section 3.2) to generate new labelled examples using the following approach. To ensure that our sampled sentences were representative of the entire corpus, we employed a stratified random sampling technique based on key metadata attributes for the documents extracted from the CELLAR portal.

We considered the distribution of sentences across 52 adoption years (1971–2022) and 20 main policy areas, resulting in 1,040 strata (52 × 20). To ensure equal representation of the strata we employed an equal allocation method which randomly sampled a fixed number of sentences from each stratum (seven sentences from each). Six of the strata contained fewer than seven sentences so they were excluded, which left 1,034 strata. Therefore, sampling seven sentences per stratum resulted in a total sample of approximately 7,200 sentences.

Thereafter, we requested the two annotators to label each sentence as either regulatory or not (a binary classification scheme). Both scholars had several years of experience applying the IGT framework for analyzing policy texts. To label such a large batch of sentences with limited resources and time, we chose to split the batch into two equal parts and have each scholar label them separately. Only in exceptionally few cases did they need to confer to arrive at an accurate label for a particular sentence.

The dataset was split randomly into an 80% training set for training the classification model and a 20% test set for measuring performance (see Section 4.1). This same test set was used to evaluate the performance of the dependency parsing approach discussed in Section 3.3.

### 3.5 XAI: explainability of LEGAL-BERT classification

The motivation for our XAI study stems from an observation made when comparing the performance of the two classification approaches presented in Sections 3.3 and 3.4. We noticed that, while both methods performed similarly in terms of accuracy scores, precision and recall, we found relatively low agreement between the two according to Krippendorff's alpha coefficient. According to Krippendorff "*It is customary to require α ≥ .800 (for statistically significant reliability). Where tentative conclusions are still acceptable, α ≥ .667 is the lowest conceivable limit*" [16]. In our comparison we observed a relatively low inter-model agreement of 0.58 (note: this value is a different measurement than those in Table 1 of Section 4.1 which were calculated against ground truth for the different methods).

This raised the hypothesis that the explanation for classifications made using the grammatical dependency approach (the rules in Section 3.3), are not necessarily transferrable to the feature-extraction transfer learning approach (Section 3.4). I.e., this approach is not purely learning the rules in Section 3.3. We wanted to discover what (potentially additional) rules or patterns the transfer learning model captures. To do this we made use of a XAI software library called DIANNA. This library implements several XAI algorithms for interpreting classification models for images, timeseries, tabular data and text. Available algorithms in DIANNA include LIME [31], RISE [29], KernelSHAP [21, 38].

We used the LIME XAI method for text data because from the models available in DIANNA, it is the only method with a clear neutral value, which is 0, in the resulting attribution values. This makes LIME's output easier to interpret. LIME is a *model agnostic*, *post-hoc*, *local* explanation method, meaning that it explains specific decisions of models, regardless of the inner workings and type of the specific model.

LIME works by *masking* (temporarily removing) random selections of tokens (words) from an input sentence and classifying the masked sentence with the to-be-explained model. This is done in an iterative manner and the effect on the classification result is monitored. A local interpretable model is then trained on these model inputs and outputs. This process generates *attribution* scores (a value between 0 and 1) for each token in the sentence. A higher attribution score indicates that the presence of that word in a particular sentence contributes more to the classification result of that sentence (regardless of the predicted class) and vice versa. As LIME masks sentences randomly, we increased the number of samples (i.e., the number of random mask variations) per sentence, until resulting attributions were stable over multiple runs.

We randomly selected 300 sentences from our test set of data and classified these sentences using our transfer learning model presented in Section 3.4. The procedure is computationally intensive and so running the XAI analysis on the entire test set of ~1,450 sentences would have been infeasible given limited resources we had to run this analysis on our computing cluster. We ran the 300 sentences through the XAI methodology explained above (to calculate the attribution scores for the sentence tokens). We retained only the TPs (true positives) and TNs (true negatives) from this data, the TPs indicating correctly classified regulatory statements and TNs indicating correctly classified non-regulatory statements.

Thereafter, for each sentence we identified the top three words with the highest attribution scores (the words most influential or indicative of the model's classification for the sentence). We sorted all words by their un-aggregated attribution scores (a specific word can appear in different sentences with different attribution scores in those contexts) and recorded the frequency of each (how often is that word highly indicative of a classification result). This was to identify which were the most commonly influential words in our evaluation (for each target class).

We also analyzed the mean *location* or mean *position* of the influential words in sentences and its effect on classification. We devised a metric for capturing this latter information by taking the start index of a word in each sentence and dividing it by the sentence length (number of characters in the sentence) and multiplying the result by one hundred.

## 4 Results

In this section we present the classification results for the two approaches described in Sections 3.3 and 3.4, as well as the XAI analysis for the model produced in Section 3.5.

### 4.1 Classification performance

The performance metrics of the classification models are shown in Table 1 below. We got the best results using Legal-BERT as a feature extraction model, just slightly better than when using either of the two other feature extraction models (base BERT and InLegal-BERT).

| Model | Accuracy | Non-regulatory | | | Regulatory | | | Krippendorff's $\alpha$* |
|---|---|---|---|---|---|---|---|---|
| | | F1 | Precision | Recall | F1 | Precision | Recall | |
| BERT | .82 | .85 | .84 | .86 | .77 | .78 | .76 | .62 |
| Legal-BERT | **.84** | **.86** | **.86** | .86 | **.79** | .80 | **.79** | **.66** |
| InLegal-BERT | .83 | **.86** | .85 | .86 | **.79** | .79 | .78 | .65 |
| Dep-parsing | .80 | .82 | .76 | **.89** | .77 | **.86** | .70 | .59 |

**Table 1.** Classification performance metrics. *Krippendorff's α values in the table are calculated by comparing the indicated model or algorithm classification result to ground truth. A separate Krippendorff's α of 0.58 was also calculated comparing the best transformer model (LegalBERT) to Dep-parsing.

The machine learning models performed slightly better than our dependency-parsing approach overall. However, we found that the dependency-parsing approach obtained the highest recall for non-regulatory statements and the highest precision for regulatory statements. This shows that the approach is stricter and more trustworthy overall in classifying a statement as regulatory. This obviously comes at the cost of recall for regulatory statements (and precision for non-regulatory statements).

As mentioned in Section 3.5, the Krippendorff's α between the top performing machine learning model and the dependency-parsing model was 0.58 (Note that Table 1 only reports Krippendorff's α values comparing the indicated models to *ground truth*), which means the agreement was relatively low between the models. Nevertheless, they both performed reasonably well (with accuracy scores above 0.8).

This observation confirms the hypothesis in Section 3.5 that the transfer learning approach is not purely learning the rules described in Section 3.3, but there are additional patterns or indicators of a regulatory statement. It is also noteworthy that the dep-parsing approach only obtained a Krippendorff score of 0.59 for agreement with ground truth from the test set, while the two LEGAL-BERT models enjoyed substantially higher agreement levels. *All source code, data and models from our study are publicly available online and documented to aid reproducibility (See Section 6 for more information).*

### 4.2 Explainability of transformer-based model

In this section, we briefly discuss the results of our XAI analysis for the feature-based transfer learning model that we developed (described in Section 3.4). Figure 5 below plots the most influential or indicative words for the regulatory class (with a cut-off frequency of at least five). The frequency values represent how often a word appears among the top three highest attribution scores for a sentence in our test set:

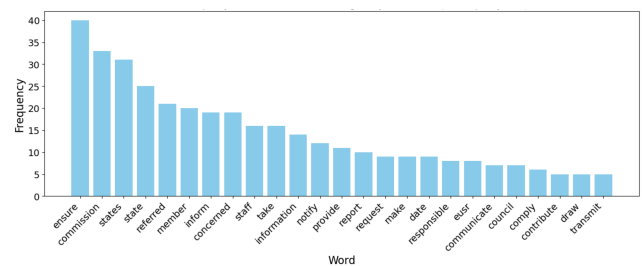

**Figure 5.** Most common words that are influential towards a regulatory classification in our XAI sample. Frequency values represent how often a word appears among the top three highest attribution scores for a regulatory classification. There is a cut-off for frequencies below five.

We plotted the same information for the true negatives in Figure 6. We observe that 60% (15 out of 25) of the words in Figure 5 play the role of verbs indicating an action that forms part of the aim of the legal rule according to the ADICO framework. In contrast, there are no verb roles in Figure 6 for the non-regulatory classifications.

If we hypothesize a pattern for Figure 6, the words collectively appear to typify diction for expressing preconditions and contextual details about EU legal rules. For example, consider the word "following" (the second most frequent indicator of a non-regulatory sentence in Figure 6). This word is often used within EU legislation to describe amendments to prior regulations. Amendments are not considered regulatory according to IGT and our domain experts because they adjust details pertaining to core legal rules. IGT is more interested in the core legal rules themselves rather than the adjustments made to them.

Consider the example excerpt: "*Implementing Decision 2011/861/EU is amended as follows…Article 2 is replaced by the following…*". Phrases such as "*Article 2 is replaced by the following*" are quite common in our training data. The phrase "replaced by the following" occurs in 275 of the 7,200 training sentences and in approximately 14,000 sentences in the EU law source data from which the training data was sampled.

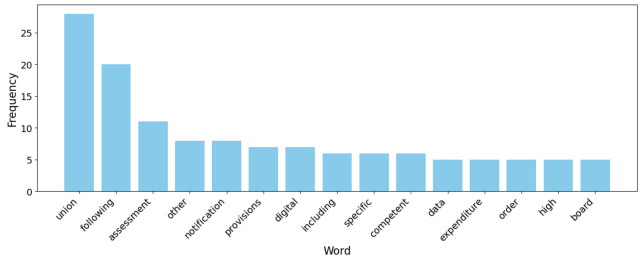

**Figure 6.** Most common words that are influential towards a non-regulatory classification in our XAI sample. Frequency values represent how often a word appears among the top three highest attribution scores for a non-regulatory classification (frequency cut-off below five).

Examining the mean position of words in a sentence, and how this affects the transformer model's classification behavior, we did not notice a significant difference between the regulatory and non-regulatory classes. Words which signal both regulatory and non-regulatory classification generally occur early in a sentence within the first 10% of characters (see Table 2).

|  | Regulatory | | Non-regulatory | |
| --- | --- | --- | --- | --- |
|  | Word pos. (%) | Sent. chars | Word pos. (%) | Sent. chars |
| Mean | 8.1 | 217 | 8.1 | 441 |
| Median | 7 | 166 | 8 | 246 |
| Std dev | 5.4 | 192 | 5 | 548 |

**Table 2.** Relative sentence position of the most influential words towards regulatory and non-regulatory classification. E.g., 8% means that the most influential words generally appear within the first 8% of characters in the sentence. "Sent. chars" represents the number of characters in the sentence.

However, a significant difference we noticed is in the length of sentences between the two classes. Non-regulatory sentences were found to be at least twice as long as regulatory ones in our *XAI sample* (see Table 2). Although our training data did not show a significant difference in mean and median sentence length between classes, the standard deviation for non-regulatory sentences was much higher than for regulatory ones (like the XAI sample). Therefore, sentence length could also be an implicitly learned feature that contributes to non-regulatory classifications. Non-regulatory sentences often being longer could be due to contextual information about legal rules requiring lengthy sentences to fully elaborate on the preconditions and criteria of the rule.

In summary, we can observe that the feature-based transfer learning model appears to have learned common verbs (forming part of the aim of a regulatory sentence) that signal regulatory classification. We can also generally observe that words commonly used to describe contextual information, preconditions and criteria for legal rules are generally indicators of a non-regulatory classification.

### 4.3 Comparison of transformer model and dep-parser

We wish to characterize the strengths and weaknesses of the two approaches in the hopes that it leads to a hybrid method to improve performance. In Section 4.1 we evaluated the performance according to standard metrics. Here, we compare the classification behaviors of the approaches by analyzing examples of successful classifications as well as misclassifications.

As seen in Table 1, the transformer model had better recall for the regulatory class. We found most of the misclassifications of regulatory statements by the dep-parser to be due to unsuccessful identification of the attribute of the legal rule. Table 3 gives some example regulatory sentences misclassified by the dep-parser approach together with the reasons for why.

| Sentence excerpt | Misclassification reason |
| --- | --- |
| "It shall inform the Council of any difficulties…" | Implicit attribute ("It") |
| "They shall keep the Head of the Union Delegation in…" | Implicit attribute ("They") |
| "The ISSB shall monitor and review…" | Unknown agent noun ("ISSB") |
| "An ARM shall continuously monitor in real-time…" | Unknown agent noun ("ARM") |

**Table 3.** Example regulatory sentences misclassified as non-regulatory by the dep-parser and correctly classified by the transformer model, with reasons for dep-parser misclassification.

In Section 3.3, the second criteria for classifying a regulatory statement requires identification of an agent noun. However, in many cases this step is difficult due to several limitations. Firstly, a regulatory sentence sometimes contains *implicit attributes,* usually pronouns such as *"It"* and *"They"* instead of explicit attributes such as *"Member States"* or *"Reporting authorities"*. Sentences with implicit attributes can either be regulatory or non-regulatory and it is non-trivial to distinguish between the two cases.

Secondly, some agent nouns such as organization name abbreviations are often unrecognized as PROPN (proper nouns and, therefore, agent nouns) by the SpaCy dependency parsing model or by our ConceptNET dictionary of agent noun phrases (see Section 3.3). For example, the International Sustainability Standards Board (ISSB) and Approved Reporting Mechanisms (ARMs) are not recognized by the dep-parser model as proper nouns, whereas terms such as UK and EU are.

We found that the transformer model, in many cases, does not have these shortcomings. I.e., it correctly classifies some regulatory sentences which the dep-parser cannot, due to the attribute identification issues mentioned above. The transformer model is therefore better at correctly dealing with implicit attributes and unknown organization names.

However, there are also some shortcomings in the transformer model's ability to learn a fully accurate representation of an attribute in a sentence. Consider the example sentence: "*The data exchange shall comply with the FLUX Vessel Position Implementation Document adopted by NEAFC*". The phrase "*The data exchange*" here does not refer to an agent noun who is the target of the legal rule, but rather a process. This sentence is therefore non-regulatory according to the IGT definition in Section 3.1. However, the transformer model classifies it as regulatory. There are over ten cases (around 1% of the test set) of this type of misclassification. The dep-parser classifies such cases correctly because it correctly identifies the nouns under consideration as not having agenthood.

To take advantage of these two strengths by the dep-parser and transformer model, a possible way to fuse them could be to classify cases where the dep-parser fails to identify an attribute using the transformer model, while handling the rest with the dep-parser. This method benefits from the increased precision of the dep-parser for identifying attributes. I.e., if it identifies an attribute then we can generally trust that it is a valid one and not one of the false positives above such as "*The data exchange*". It also benefits from the transformers superior ability to identify implicit attributes and unseen agent nouns.

## 5 Challenges and limitations

We encountered several challenges in this work which are notable for the reader and provide interesting avenues for future work. We discuss the most significant ones below.

*Attribute identification:* for the dependency parsing approach there were many false negative cases which revealed complexity and variance in where the attribute of a regulatory sentence can be placed. In our implementation of this approach, we unfortunately could not account for all possible patterns due to the complexity of the grammatic structures that govern how this can be expressed. Given the good performance of the first version of our dependency parser, it implies that such grammatic variations occur relatively infrequently in EU Law. However, we plan to extend this implementation with several new patterns based on exploiting finer-grained dependency types and paths (e.g., the "agent" dependency) which aid attribute identification.

*Optical Character Recognition (OCR) issue*: for some legislation before 1990, the full texts were only available in PDF documents run through optical character recognition (OCR) software to extract the text. Due to the double-column layout in some of these documents, there were some errors made by the software. Specifically, when page breaks occurred and sentences were spread across multiple pages, the OCR software incorrectly reordered and spliced parts of different sentences together. This created some nonsensical sentences.

> To eliminate such sentences (because they were not possible to detect and remove automatically), we had to sample sentences for the evaluation phase primarily from the years post 1990s in the dataset (from 2000 to 2023) which were not subject to this anomaly. Excluding these years from consideration obviously affected the generalizability of our results. However, human analysis of the language and phrasing used before 2000 in the regulations showed no significant differences and we therefore expect our results to generalize to pre-1990s documents.

*XAI methods:* state-of-the-art XAI methods for text classification based on masking of tokens can currently only assign attribution scores to individual tokens (words). We hypothesize that the assignment of attribution scores to phrases (*n-grams* or consecutive combinations of words) in sentences are likely to be more interpretable by human experts.

Unfortunately, current methods do not assign attribution scores to n-grams, nor do they currently implement n-gram masking procedures as discussed in Section 3.5. For example, consider the regulatory sentence "*Citizens must separate their recyclables.*". Suppose that the words "*Citizens*", "*must*" and "*separate*" obtain attribution scores of 0.55 each. And suppose the words "*their*" and "*recyclables*" obtain scores of 0.21. One can easily observe that the first three words are more relevant for classifying the sentence as regulatory with higher scores. However, with this approach: 1) one cannot distinguish between the importance of each of the first three words since they have the same score, and 2) one cannot infer if there is a specific combination of these words which is a clearer indicator of the classification result.

The theory is that if we could assign attribution scores to phrases e.g., "*Citizens must separate*" then we might be able to obtain a higher attribution score than 0.55 for this combined phrase than the individual words. And this would lead to classification results that are more interpretable by domain experts. Conceptually, such a feature is not difficult to implement and could enhance the explainability of classification results.

## 6 Reproducibility

To help researchers reproduce and improve upon this work, we have made our code, analysis and data publicly available online. We created a Github organization called nature-of-eu-rules (https://github.com/nature-of-eu-rules) which contains all the code we used in this study. Three of the four code repositories in the organization are used in this study. The code for extracting full text regulation documents and associated metadata from EURLEX, is located in the data-extraction[8] repository. To process these documents and extract the potentially regulatory sentences, we used code in the data-preprocessing[9] repository. Finally, the code for our two methods of classifying regulatory sentences is in the regulatory-statement-classification[10] repository.

We have made our raw text dataset of regulatory documents available on the Zenodo digital resource archival platform here: https://doi.org/10.5281/zenodo.8174176. All processed data used for training, evaluation and analysis is available also on Zenodo here: https://doi.org/10.5281/zenodo.12760951. The code and data repositories have further detailed information about how to interpret the files and contents.

## 7 Conclusions and future work

We have presented two approaches for classifying sentences in EU law as either regulatory or not. The first approach analyses the grammatical dependency parse trees (generated by pretrained neural network-based models) of the sentences. The second is based on a feature-based transfer learning model (based on LEGAL-BERT) evaluated on the largest corpus of human-labelled sentences for binary text classification of regulatory statements. Both models obtained F1-scores exceeding 0.8 with a relatively low level of agreement according to the Krippendorff alpha metric. We also identified a concrete way to combine strengths of both approaches which could significantly improve performance. *All source code, data and models generated and used in our study are publicly available online and documented to aid reproducibility; links will be disclosed after the review process.*

---
[8] https://github.com/nature-of-eu-rules/data-extraction
[9] https://github.com/nature-of-eu-rules/data-preprocessing
[10] https://github.com/nature-of-eu-rules/regulatory-statement-classification

While we obtained good performance even with a preliminary dependency parsing approach, incorrectly classified sentences were due to complexity and variance in grammatic phrasing for representing regulatory sentences. Our dependency rules could be improved to deal with such issues by distinguishing between dependency types and paths. A promising finding, though, is that such cases occur relatively infrequently in EU Law.

Preliminary analysis of the classification results of the feature-based transfer learning model, using state-of-the-art XAI methods, reveals that the presence of common verbs (forming part of an aim for a legal rule according to the IGT framework) in a sentence are strong indicators of a regulatory classification result. For a non-regulatory classification result, words typically describing contextual information, preconditions and criteria for legal rules appear to be indicators.

We aim to also improve our XAI method by implementing n-gram masks of varying lengths to identify phrases responsible for classification results. Recently, SHAP has been argued to have advantages over the LIME method used in this study, potentially making it more suitable for use in legal contexts [38]. Applying SHAP analysis of our classifications could prove interesting to determine if it improves interpretability of the model since it provides both local and global interpretability (ability to aggregate explanations for individual predictions).